\documentclass{article}

\usepackage{microtype}
\usepackage{graphicx}
\usepackage{subcaption}
\usepackage{booktabs} 

\usepackage{hyperref}

\usepackage[accepted]{icml2026/icml2026}

\usepackage{amsmath}
\usepackage{amssymb}
\usepackage{mathtools}
\usepackage{amsthm}
\usepackage{multirow}

\usepackage[capitalize,noabbrev]{cleveref}

\theoremstyle{plain}

\theoremstyle{definition}

\theoremstyle{remark}

\icmltitlerunning{Aligning Confidence with Correctness for LLM Error Detection}

\makeatletter
\renewcommand{\ICML@appearing}{%
  \textit{Preliminary work. Under review.}
}
\makeatother

\begin{document}

\twocolumn[
  \icmltitle{Know When You're Wrong: Aligning Confidence with Correctness for LLM Error Detection}
%


  \begin{icmlauthorlist}
    \icmlauthor{Xiaohu Xie}{amzn}
    \icmlauthor{Xiaohu Liu}{amzn}
    \icmlauthor{Benjamin Yao}{amzn}
  \end{icmlauthorlist}

  \icmlaffiliation{amzn}{Alexa AI, Amazon, Seattle, WA, USA}

  \icmlcorrespondingauthor{Xiaohu Xie}{xxie@amazon.com}
  \icmlcorrespondingauthor{Xiaohu Liu}{derecliu@amazon.com}
  \icmlcorrespondingauthor{Benjamin Yao}{benjamy@amazon.com}

  \icmlkeywords{Machine Learning, ICML}

  \vskip 0.3in
]


\printAffiliationsAndNotice{}  

\begin{abstract}
As large language models (LLMs) are increasingly deployed in critical decision-making systems, the lack of reliable methods to measure their uncertainty presents a fundamental trustworthiness risk.
We introduce a normalized confidence score based on output anchor token probabilities: classification labels for structured tasks and self-evaluation responses (Yes/No) for open-ended generation.
This enables direct detection of errors and hallucinations with minimal overhead and without external validation.
We make three key contributions.
First, we propose a normalized confidence score and self-evaluation framework that exposes reliable confidence estimates for error detection across seven diverse benchmark tasks and five LLMs of varying architectures and sizes.
Second, our theoretical analysis reveals that supervised fine-tuning (SFT) yields well-calibrated confidence through maximum-likelihood estimation,
whereas reinforcement learning methods (PPO, GRPO) and DPO induce overconfidence via reward exploitation.
Third, we propose post-RL SFT with self-distillation to restore confidence reliability in RL-trained models.
Empirical results demonstrated that SFT improved average confidence-correctness AUROC from 0.806 to 0.879 and reduced calibration error from 0.163 to 0.034 on Qwen3-4B, while GRPO and DPO degraded confidence reliability.
We demonstrated practical value through adaptive retrieval-augmented generation (RAG) that selectively retrieves context when the model lacks confidence, using only 58\% of retrieval operations to recover 95\% of the maximum achievable accuracy gain on TriviaQA.
\end{abstract}

\section{Introduction}

Large language models frequently generate plausible but incorrect outputs with unwarranted confidence—a phenomenon commonly termed ``hallucinations''~\cite{ji2023survey,maynez2020faithfulness,zhang2023siren}.
This tendency to present unreliable predictions as certain undermines deployment in critical decision-making systems, from healthcare diagnostics to financial advising.
The fundamental challenge lies not in eliminating errors entirely, but in enabling models to reliably quantify their uncertainty and signal when predictions may be unreliable.

We address this challenge by extracting confidence estimates directly from model output probabilities, calibrating them to match actual performance.
In high-stakes domains, well-calibrated confidence estimates prevent costly errors, maintain user trust~\cite{ji2023survey,barkan2025llm}, and enable selective application of expensive strategies—retrieving context for uncertain queries, engaging deeper reasoning for complex problems, or escalating to human experts~\cite{dhuliawala2024chain,advani2026right,joren2024sufficient}—only when needed.

Existing approaches have limitations.
Detection methods like self-consistency~\cite{wang2022self} sample multiple outputs but lack quantitative confidence measures, while correction methods like Chain-of-Verification~\cite{dhuliawala2024chain} fix hallucinations uniformly without identifying which outputs need correction.
Our probability-based approach provides immediate, calibrated confidence estimates enabling selective intervention.

\textbf{Contributions.}
We investigate whether LLM output probabilities can serve as reliable confidence indicators for error detection, demonstrating that probability-derived confidence strongly correlates with prediction accuracy.
We extend this to general generation tasks via self-evaluation, which converts free-form generation into binary classification by prompting models to assess their own answers.
Unlike text-based detection methods that require costly multiple sampling (especially problematic for RL models with sharpened distributions), our probability-based approach provides numerical confidence values enabling precise threshold-based decision making with a single forward pass.
We provide theoretical analysis of how training paradigms affect calibration: pre-training and SFT naturally promote calibration through maximum-likelihood estimation, 
while DPO and trust-region RL methods, like PPO and GRPO, degrade it by exploiting rewards through advantage-weighted gradients.
Through post-RL SFT, we restore calibration across 7 benchmark tasks on Qwen3-4B.

We validate practical value through adaptive RAG, demonstrating that well-calibrated confidence enables selective retrieval that captures most accuracy gains at a fraction of the cost.

\section{Related Work}
\label{sec:related}

\textbf{Hallucination Detection in LLMs.}
Detecting hallucinations and factual errors has become critical~\cite{ji2023survey,zhang2023siren}. Self-consistency sampling~\cite{wang2022self} generates multiple outputs to identify likely correct answers, while Chain-of-Verification~\cite{dhuliawala2024chain} generates verification questions to check claims. 
However, these methods require significant overhead—multiple forward passes for self-consistency, and additional generation for verification. Our probability-based approach provides immediate confidence estimates without extra generation, enabling efficient error detection.

\textbf{Self-Evaluation and Uncertainty Quantification.}
Recent work investigates LLMs' ability to evaluate their own correctness. 
\cite{kadavath2022language} showed that language models can evaluate the validity of their own claims by predicting P(True) using probabilities of all tokens in a response. 
\cite{kuhn2023semantic} introduced semantic entropy, which incorporates linguistic invariances by clustering semantically equivalent answers to measure uncertainty. 
\cite{barkan2025llm} revealed that self-evaluation remains poorly developed, with models often exhibiting poor calibration. 

Our work differs from \cite{kadavath2022language} in three key aspects. 
First, unlike their use of raw P(True) probabilities, we introduce \textit{normalized confidence} that accounts for the constrained output space by normalizing over all valid answer options (Equation~\ref{eq:confidence}). 
This normalization substantially improves discriminative power, with AUROC (Area Under the Receiver Operating Characteristic curve) gains up to +33.1\% (see Appendix, Table~\ref{tab:normalized_vs_unnormalized}).
Second, while \cite{kadavath2022language} relies on few-shot prompting and shows calibration significantly improves with more examples, our method achieves strong zero-shot performance without requiring few-shot demonstrations, making it more practical for deployment.
Third, \cite{kadavath2022language} does not investigate how different training paradigms (SFT vs. RL vs. DPO) affect calibration quality or provide methods to restore calibration. 
\cite{kuhn2023semantic} requires generating multiple diverse answers to cluster semantically equivalent responses, adding computational overhead. 

\textbf{Training Methods and Calibration.}
Different training paradigms significantly impact model calibration. \cite{tian2023calibration} showed that RLHF causes systematic overconfidence, while \cite{leng2024taming} explored reward calibration techniques to reduce it. 
However, these works focus on empirical observations without explaining the mechanistic causes or comparing different alignment methods (e.g., trust-region RL vs. DPO). 
We extend this line of work by analyzing why advantage-weighted gradients cause miscalibration and comparing calibration properties across multiple training paradigms including SFT, PPO, GRPO, and DPO (see Section \ref{sec:calibration_analysis}).

\textbf{Adaptive Computation in LLMs.}
Several works explore adaptive computation for efficiency. \cite{jeong2024adaptive} proposed Adaptive-RAG for selective retrieval, while \cite{asai2023self} introduced Self-RAG using reflection tokens. However, these approaches rely on models generating explicit text-based decisions, which can be unreliable for RL-trained models. Our probability-based confidence enables robust threshold-based decision making, as demonstrated through our adaptive applications.
\section{Know When You're Wrong}
\label{sec:method}

We propose a framework for enabling LLMs to reliably signal uncertainty by extracting confidence estimates from output probabilities. We present three key components: 
(1) a normalized confidence score for classification tasks, 
(2) a self-evaluation framework for open-ended generation, and 
(3) evaluation metrics for discriminative power and calibration quality. 
This methodology enables LLMs to know when they don't know, providing the foundation for adaptive systems.

\subsection{Classification Tasks}
\label{sec:classification}

Given input $x$, an LLM $\pi_\theta$ generates output $y = (y_1, \ldots, y_T)$ by sequentially predicting tokens, producing probability distribution $\pi_\theta(y_t | x, y_{<t})$ over the vocabulary for each token $y_t$. 
We define the raw confidence as the product of token probabilities:
\begin{equation*}
c(y|x) = \prod_{t=1}^{T} \pi_\theta(y_t | x, y_{<t})
\end{equation*}
For classification, the model selects label $y$ from predefined classes $\mathcal{Y}$. 
To account for this constraint, we define normalized confidence:
\begin{equation}
\label{eq:confidence}
\hat{c}(y|x) = \frac{c(y|x)}{\sum_{y'\in \mathcal{Y}}c(y'|x)}.
\end{equation}

Normalized confidence is more robust as it considers the constrained output space, instead of all possible tokens.
As demonstrated in Appendix (Table~\ref{tab:normalized_vs_unnormalized}), normalized confidence consistently outperforms raw confidence across all evaluated models and tasks, with AUROC improvements up to +33.1\% on classification tasks.

\subsection{Self-Evaluation for General Tasks}
\label{sec:selfeval}

For generation tasks (e.g., math, reading comprehension), answer probabilities cannot serve as confidence since the output space is large and diverse.
We propose self-evaluation to convert open-ended tasks into binary classification:
\begin{enumerate}
    \item Generate answer: $\hat{y} = \text{decode}(\pi_\theta(\cdot|x))$
    \item Use self-evaluation prompt:
    \begin{equation*}s=\text{Is this answer correct? Answer only Yes/No.}\end{equation*}
    This yields confidence distribution $c_s(t)=\pi_\theta(t|x,\hat{y},s)$ for $t\in\{\text{Yes},\text{No}\}$
    \item Similar to Equation~\ref{eq:confidence}, calculate normalized self-eval confidence:
    \begin{equation}
    \label{eq:selfeval_confidence}
    \hat{c}_s = \frac{c_s(\text{Yes})}{c_s(\text{Yes}) + c_s(\text{No})}
    \end{equation}
\end{enumerate}
Appendix~\ref{sec:selfeval_example} illustrates this process with a factual question.

\textbf{Implementation:} Both "Yes" and "No" are single tokens in standard tokenizers. We examine only the probability distribution over the first token position without generating multi-token responses. Using vllm with \texttt{max\_tokens=1} and \texttt{n\_log\_probs=20}, we retrieve the top-20 tokens by probability and default $\pi_\theta(t|x,\hat{y},s)=0$ for $t \notin \text{top-20}$. This single-token approach adds minimal overhead while providing reliable confidence estimates.

\subsection{Confidence Evaluation}
\label{sec:evaluation}

We evaluate on 7 tasks spanning classification (BoolQ~\cite{clark2019boolq}, AG News~\cite{zhang2015character}, CommonsenseQA~\cite{talmor2019commonsenseqa}, HellaSwag~\cite{zellers2019hellaswag}) and generation (GSM8K~\cite{cobbe2021training}, TriviaQA~\cite{joshi2017triviaqa} with/without context). Classification tasks use normalized confidence based on output probabilities; generation tasks use self-evaluation. See Appendix~\ref{sec:benchmark_details} for task descriptions.

\begin{table*}[t]
\centering
\small
\begin{tabular}{l|l|cccc|ccc|c}
\toprule
\textbf{Model} & \textbf{Metric} & \textbf{BoolQ} & \textbf{CQA} & \textbf{AG News} & \textbf{HellaSwag} & \textbf{GSM8K} & \textbf{TriviaQA} & \textbf{TriviaQA-RC} & \textbf{Average} \\
\midrule
\multirow{3}{*}{\shortstack{Qwen3\\4B}} & Acc. \% & 84.40 & 78.21 & 85.63 & 79.64 & 92.49 & 54.21 & 72.71 & \textbf{78.18} \\
 & AUROC & 0.817 & 0.803 & 0.750 & 0.809 & 0.773 & 0.829 & 0.863 & \textbf{0.806} \\
 & ECE & 0.149 & 0.196 & 0.132 & 0.182 & 0.069 & 0.255 & 0.158 & \textbf{0.163} \\
\midrule
\multirow{3}{*}{\shortstack{Qwen3\\30B}} & Acc. \% & 89.11 & 85.09 & 83.54 & 87.36 & 96.36 & 72.99 & 83.08 & \textbf{85.36} \\
 & AUROC & 0.778 & 0.847 & 0.710 & 0.857 & 0.787 & 0.859 & 0.835 & \textbf{0.810} \\
 & ECE & 0.100 & 0.124 & 0.155 & 0.102 & 0.026 & 0.181 & 0.124 & \textbf{0.116} \\
\midrule
\multirow{3}{*}{\shortstack{Gemma3\\4B}} & Acc. \% & 83.94 & 69.70 & 77.80 & 63.10 & 86.05 & 53.36 & 70.47 & \textbf{72.06} \\
 & AUROC & 0.788 & 0.756 & 0.718 & 0.698 & 0.686 & 0.775 & 0.819 & \textbf{0.749} \\
 & ECE & 0.154 & 0.295 & 0.211 & 0.357 & 0.121 & 0.411 & 0.244 & \textbf{0.256} \\
\midrule
\multirow{3}{*}{\shortstack{Gemma3\\12B}} & Acc. \% & 86.48 & 78.71 & 81.55 & 79.97 & 92.87 & 72.04 & 80.42 & \textbf{81.72} \\
 & AUROC & 0.747 & 0.820 & 0.683 & 0.804 & 0.801 & 0.814 & 0.831 & \textbf{0.786} \\
 & ECE & 0.129 & 0.200 & 0.179 & 0.181 & 0.045 & 0.231 & 0.148 & \textbf{0.159} \\
\midrule
\multirow{3}{*}{\shortstack{GLM4\\9B}} & Acc. \% & 88.26 & 81.82 & 86.21 & 83.26 & 65.43 & 60.52 & 75.43 & \textbf{77.28} \\
 & AUROC & 0.826 & 0.833 & 0.809 & 0.827 & 0.734 & 0.748 & 0.768 & \textbf{0.792} \\
 & ECE & 0.037 & 0.025 & 0.086 & 0.016 & 0.137 & 0.315 & 0.164 & \textbf{0.111} \\
\bottomrule
\end{tabular}
\caption{Performance across open-source models using normalized confidence. Classification tasks use output probabilities; Generation tasks use self-evaluation. CQA = CommonsenseQA.}
\label{tab:public_model_performance}
\end{table*}

We use AUROC to measure the discriminative power for confidence-based error detection.
AUROC quantifies how well confidence scores can distinguish between correct and incorrect predictions: an AUROC of 1.0 indicates perfect separation (all correct predictions have higher confidence than incorrect ones), while 0.5 indicates random guessing.
We use Expected Calibration Error (ECE) to quantify miscalibration by measuring the gap between predicted confidence and empirical accuracy.
To calculate ECE, we partition samples from each task into 10 equal-mass bins based on confidence and measure the average absolute difference between confidence and accuracy for each bin:
\begin{equation*}
\text{ECE} = \sum_{b=1\dots 10} \frac{|B_b|}{N} |\text{acc}(B_b) - \text{conf}(B_b)|
\end{equation*}
where $B_b$ is the set of samples in bin $b$, and $N$ is the total number of samples.

We evaluate our approach on several state-of-the-art open-source language models with diverse architectures: Qwen3~\cite{qwen3_2025} with a 4B dense model and a 30B Mixture-of-Experts (MoE) model, Gemma-3~\cite{team2023gemma} with dense architectures at 4B and 12B parameters, and GLM-4~\cite{glm2024} (9B parameters). 
These models represent diverse architectures and training approaches, allowing us to assess the generalizability of our confidence-based error detection method.
Table~\ref{tab:public_model_performance} presents the evaluation results across both classification and self-evaluation tasks.
For classification tasks, we use output probabilities directly; for self-evaluation tasks, we prompt models to assess their own answers.
The results demonstrated that model confidence exhibited strong discriminative power for prediction correctness across various model families. 
Performance remained consistent across both classification tasks (BoolQ, CommonsenseQA, AG News, HellaSwag) and self-evaluation tasks (GSM8K, TriviaQA, TriviaQA-RC), demonstrating the versatility of our confidence-based approach.
However, these models exhibit suboptimal confidence calibration with high ECE despite strong discriminative power.
Investigation reveals that the miscalibration consistently manifests as \textit{distribution sharpening} across various models. 
The confidence collapses toward zero when accuracy is low, and sharply jumps to near one when accuracy crosses a certain threshold.
Figure~\ref{fig:calibration_curves} visualizes this pattern for Qwen3-4B-Instruct (the baseline curve), with detailed per-bin results in Appendix Table~\ref{tab:qwen3_4b_norm_bins}.
Note that nearly all modern LLMs go through reinforcement learning as the last training stage.
This distribution sharpening phenomenon provides the empirical motivation for our theoretical analysis in Section~\ref{sec:calibration_analysis}, where we explain why RL fundamentally induces this behavior through reward exploitation.
\section{Calibration Analysis}
\label{sec:calibration_analysis}
The confidence evaluation results demonstrate that model confidence exhibits strong discriminative power for error detection. 
However, significant miscalibration remains, particularly in models trained with reinforcement learning.
In this section, we provide theoretical analysis of the source of this miscalibration.

Consider a general task with data distribution $P_{\text{data}}(y|x)$ where $x$ is the input and $y$ is the expected output.
For a well-calibrated model $\pi_\theta(y|x)$, we expect $\pi_\theta(y|x) \approx P_{\text{data}}(y|x)$, meaning the KL divergence between $P_{\text{data}}$ and $\pi_\theta$ should be minimal: $D_{\text{KL}}(P_{\text{data}} \| \pi_\theta) \approx 0$.

\subsection{Pre-training and SFT}

Pre-training (PT) and Supervised Fine-Tuning (SFT) use next token prediction as the training objective. Given a sequence of tokens $y = (y_1, \ldots, y_T)$ and input context $x$, the model is trained to minimize the cross-entropy loss:
\begin{equation*}
\mathcal{L}_{\text{CE}} = -\sum_{t=1}^{T} \log \pi_\theta(y_t | x, y_{<t})
\end{equation*}
where $\pi_\theta$ denotes the model's predicted distribution parameterized by $\theta$. This cross-entropy loss is equivalent to minimizing the KL divergence between the data distribution $P_{\text{data}}$ and the model distribution $\pi_\theta$:
\begin{equation*}
\begin{split}
\mathcal{L}_{\text{CE}} &= \mathbb{E}_{(x,y) \sim P_{\text{data}}}[-\log \pi_\theta(y|x)] \\
&= D_{\text{KL}}(P_{\text{data}} \| \pi_\theta) + \text{const}
\end{split}
\end{equation*}

Thus, minimizing cross-entropy loss performs maximum-likelihood estimation (MLE), which directly learns to match the data distribution. This MLE objective naturally promotes calibration: the model's output probabilities reflect the empirical frequencies in the training data. When the training data is representative of the true task distribution, the model learns well-calibrated confidence estimates.

\subsection{Reinforcement Learning}

Trust-region policy optimization methods like PPO (Proximal Policy Optimization) and GRPO (Group Relative Policy Optimization) are the most widely adopted RL algorithms for LLM training. These methods, inspired by TRPO (Trust Region Policy Optimization), share the common characteristic of constraining policy updates.
For simplicity, RL in this paper specifically refers to trust-region RL methods.
In contrast to SFT, these methods optimize a fundamentally different objective using advantage-weighted policy gradients. The PPO objective is:
\begin{equation*}
\mathcal{L}_{\text{PPO}}(\theta) = \mathbb{E}_t \left[ \text{clip}_\epsilon(\frac{\pi_\theta(y_t|x, y_{<t})}{\pi_{\theta_{\text{old}}}(y_t|x, y_{<t})}) \hat{A}_t \right]
\end{equation*}
where $\text{clip}_\epsilon$ is a function to clip this ratio to a trust region, and $\hat{A}_t$ is the advantage estimate computed from the reward function.

Critically, this objective:
\begin{enumerate}
\item \textbf{Optimizes for reward, not data likelihood}: The model learns to maximize expected rewards rather than to match the data distribution. Actions (tokens) with higher advantages receive higher probability mass regardless of their frequency in the data.

\item \textbf{Sharpens distributions through reward exploitation}: The advantage weighting $\hat{A}_t$ causes the model to concentrate probability mass on high-reward actions. 
Through repeated updates, actions with even slight advantages ($\hat{A}_t > 0$) receive exponentially more probability mass, eventually approaching near-certainty. 
For an RL model to be well-calibrated with respect to its reward-based training objective, it would need to assign probabilities that reflect the relative advantage values. 
However, this sharpening effect causes severe miscalibration: the model becomes overconfident even when advantages are small.

\item \textbf{Does not perform MLE}: Unlike SFT, RL does not minimize $D_{\text{KL}}(P_{\text{data}} \| \pi_\theta)$. The resulting policy $\pi_\theta$ is not a maximum-likelihood estimate of the data distribution. While a well-calibrated SFT model assigns probabilities proportional to empirical frequencies in training data, RL models assign probabilities to maximize expected rewards.
\end{enumerate}

This fundamental difference explains why RL-trained models exhibit poorer calibration: their confidence estimates reflect reward optimization rather than data distribution matching. 
The distribution sharpening inherent to RL training is fundamentally at odds with the goal of maintaining calibrated confidence that reflects true prediction uncertainty.
Recent work by \cite{matsutani2025rl} provides complementary evidence from a reasoning path perspective, showing that RL ``squeezes'' reasoning into concentrated pathways while SFT ``expands'' it across diverse steps, further supporting the mechanistic differences between these training paradigms and their impact on model behavior.

\subsection{Direct Preference Optimization}

Direct Preference Optimization (DPO)~\cite{rafailov2023direct} represents an alternative Preference Learning approach.
Unlike TRPO-based methods, DPO optimizes a preference-based objective using maximum-likelihood estimation. Given preferred output $y_w$ and less preferred $y_l$, DPO minimizes:
\begin{equation*}
\mathcal{L}_{\text{DPO}} = -\mathbb{E}_{(x, y_w, y_l) \sim D} \log \sigma \Big( \beta \log \frac{r(y_w)}{r(y_l)}\Big)
\end{equation*}
where $r(y) = \pi_\theta(y | x)/\pi_{\text{ref}}(y | x)$ is the probability ratio of a specific response relative to the starting point.

While DPO uses maximum-likelihood estimation, it optimizes a fundamentally different objective than SFT. 
DPO performs MLE over \textit{preference probabilities} $\sigma(\beta \log r(y_w)/r(y_l))$---the probability that $y_w$ is preferred over $y_l$---rather than \textit{output probabilities} $\pi(y|x)$. 
This distinction has critical implications for calibration:

\begin{enumerate}
\item \textbf{Optimizes preferences, not output likelihoods}: 
DPO learns to predict preference rankings, not the probability of generating specific tokens. 
The model is trained to maximize the gap between $r(y_w)$ and $r(y_l)$, not to match the empirical frequency of tokens in the data.

\item \textbf{Relative vs. absolute probabilities}: 
DPO only cares about the relative improvement over the reference model (the ratio $r(y_w)/r(y_l)$), not the absolute probability $\pi(y|x)$ of any output. 
To achieve high preference accuracy, the model needs to actively push toward overconfidence to preferred outputs $y_w$, leading to miscalibrated absolute probabilities.
This mechanism drives distribution sharpening similar to trust-region RL.
\end{enumerate}

Like RL methods, DPO does not minimize $D_{\text{KL}}(P_{\text{data}} \| \pi_\theta)$, so the resulting $\pi_\theta(y|x)$ is not a maximum-likelihood estimate of the data distribution. 
While DPO avoids the advantage-weighted gradients of trust-region RL, its preference-based objective with sigmoid saturation still leads to overconfident output probabilities that reflect reward optimization rather than calibrated uncertainty.

\subsection{Empirical Validation}
To validate our theoretical analysis, we conduct controlled experiments by fine-tuning Qwen3-4B-Instruct using three distinct training paradigms—SFT, RL (GRPO), and DPO—on identical training data covering all seven benchmark tasks and their self-evaluation variants.
Notably, since Qwen3 models already undergo GRPO RL training as their final training step~\cite{qwen3_2025}, 
our experiments demonstrate that post-RL SFT can restore well-calibrated confidence in models that have been degraded by RL training, directly confirming our theoretical predictions from Section~\ref{sec:calibration_analysis}.

\begin{figure*}[t]
\centering
\includegraphics[width=0.95\textwidth]{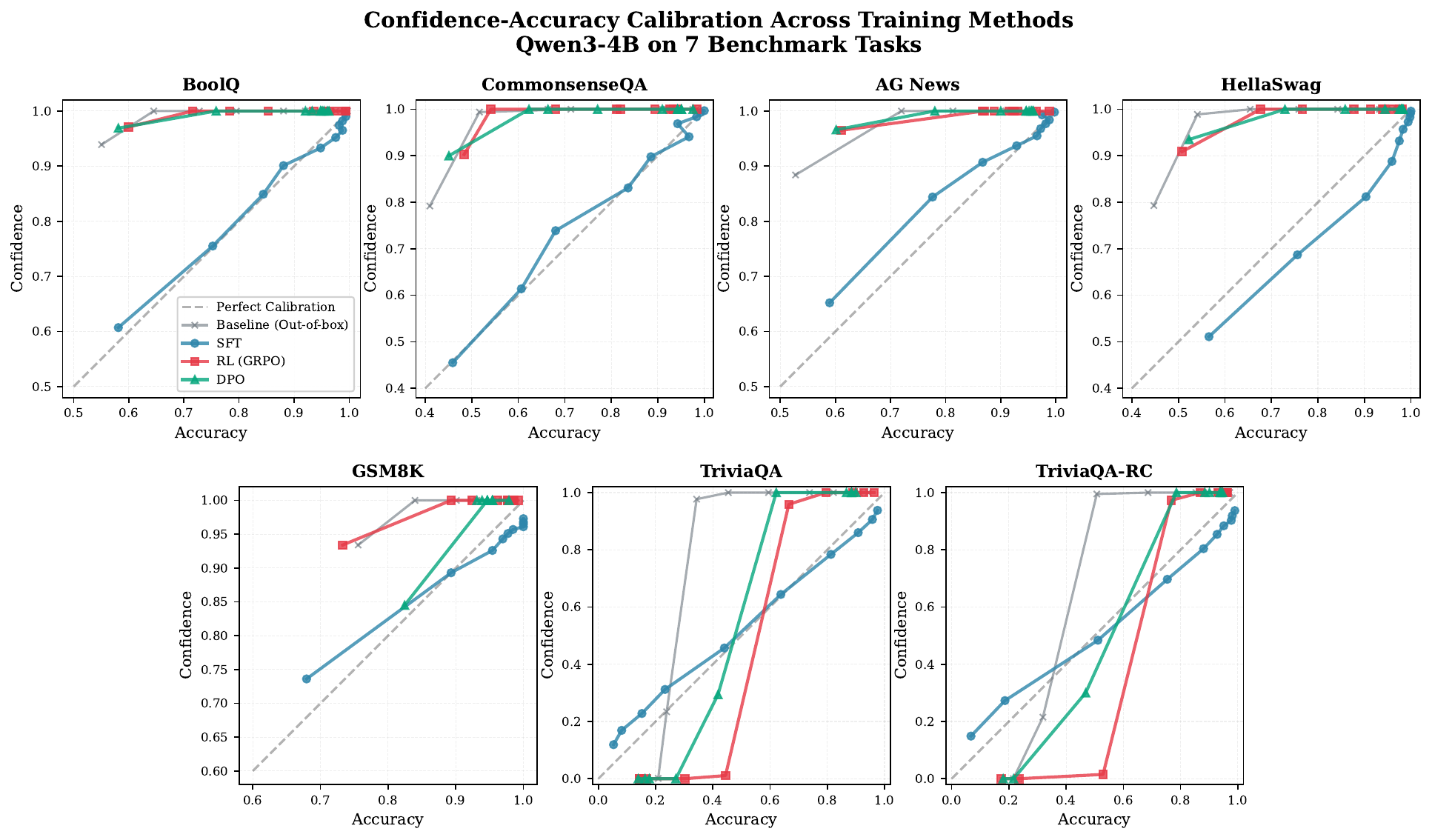}
\caption{Confidence-accuracy calibration curves comparing: Baseline (Qwen3-4B-Instruct out-of-box), SFT, RL (GRPO), and DPO training methods across 7 benchmark tasks. 
Each subplot shows the relationship between accuracy (x-axis) and confidence (y-axis) across 10 equal-mass confidence bins. 
The diagonal line represents perfect calibration. 
SFT naturally produces calibrated confidence, while RL and DPO's reward-based objectives induce distribution sharpening.}
\label{fig:calibration_curves}
\end{figure*}

For generation tasks (GSM8K and TriviaQA), we use self-distillation to generate training labels, where the model generates its own reasoning traces and we select correct ones for model training.
\textit{Self-distillation is critical for maintaining model performance.} 
By sampling from the model's own outputs rather than using original dataset labels, self-distillation minimizes disruption to the reasoning and response style, effectively preserving the model's behavior and existing capabilities.
Without self-distillation, fine-tuning Qwen3-4B on the original GSM8K dataset causes accuracy to drop from 92.5\% to 83.09\%, while self-distilled labels achieve 94.62\% accuracy.
Figure~\ref{fig:calibration_curves} visualizes the calibration patterns across all methods, with Table~\ref{tab:sft_vs_rl_summary} summarizing the key metrics.
Complete training details, including self-distillation, hyperparameters, and reward functions, are provided in Appendix~\ref{sec:training_details}.
Detailed per-bin calibration analysis is provided in Appendix Table~\ref{tab:qwen3_4b_norm_bins}.

\textbf{SFT Results.}
SFT naturally produces well-calibrated confidence estimates through its maximum-likelihood objective, which learns to match the data distribution. 
The SFT model achieves average AUROC of 0.879 and ECE of 0.034, providing both strong discriminative power and excellent calibration.
This makes SFT the preferred training paradigm when reliable uncertainty quantification is essential.

\textbf{RL/DPO Results.}
In contrast, while RL training improves task accuracy, it maintains only comparable AUROC (0.809 vs. 0.806 baseline)—RL did not improve the model's ability to discriminate between correct and incorrect predictions.
The RL model shows slightly reduced ECE (0.135 vs. 0.163 baseline).
RL improved model accuracy, which mechanically reduced the confidence-accuracy gap without addressing the underlying calibration issue.

The DPO model exhibits a similar pattern as RL. DPO ECE drops due to the accuracy improvement while AUROC drops to 0.785, indicating an even weaker confidence-accuracy alignment.
While the confidence is still sharpened, it is interesting to see that RL and DPO ``calibrate'' the threshold where the confidence jumps when accuracy is near 50\% for TriviaQA.

\begin{table*}[t]
\centering
\small
\begin{tabular}{l|l|cccc|ccc|c}
\toprule
\textbf{Method} & \textbf{Metric} & \textbf{BoolQ} & \textbf{CQA} & \textbf{AG News} & \textbf{HellaSwag} & \textbf{GSM8K} & \textbf{TriviaQA} & \textbf{TriviaQA-RC} & \textbf{Average} \\
\midrule
\multirow{3}{*}{SFT} & Acc. \% & 89.33 & 83.54 & 90.42 & 91.34 & 94.62 & 52.52 & 72.25 & \textbf{82.00} \\
 & AUROC & 0.839 & 0.847 & 0.842 & 0.898 & 0.897 & 0.908 & 0.924 & \textbf{0.879} \\
 & ECE & 0.013 & 0.014 & 0.022 & 0.040 & 0.030 & 0.050 & 0.066 & \textbf{0.034} \\
\midrule
\multirow{3}{*}{\shortstack{RL\\(GRPO)}} & Acc. \% & 87.74 & 80.18 & 88.80 & 85.42 & 94.16 & 54.43 & 73.36 & \textbf{80.58} \\
 & AUROC & 0.819 & 0.786 & 0.700 & 0.807 & 0.799 & 0.869 & 0.886 & \textbf{0.809} \\
 & ECE & 0.120 & 0.189 & 0.108 & 0.137 & 0.052 & 0.191 & 0.150 & \textbf{0.135} \\
\midrule
\multirow{3}{*}{DPO} & Acc. \% & 89.17 & 81.82 & 89.53 & 89.35 & 93.78 & 53.28 & 72.02 & \textbf{81.28} \\
 & AUROC & 0.761 & 0.803 & 0.751 & 0.850 & 0.629 & 0.851 & 0.850 & \textbf{0.785} \\
 & ECE & 0.105 & 0.172 & 0.101 & 0.100 & 0.047 & 0.172 & 0.123 & \textbf{0.117} \\
\bottomrule
\end{tabular}
\caption{Calibration comparison: SFT vs. RL (GRPO) vs. DPO training on Qwen3-4B with identical data. SFT produces well-calibrated confidence through maximum-likelihood estimation; RL and DPO lead to miscalibration through reward exploitation.}
\label{tab:sft_vs_rl_summary}
\end{table*}

\subsection{Implications for LLM Training}

Our experiments reveal that self-evaluation emerges as a capability in many current open-source LLMs—models can assess their own output correctness through simple prompting, as evidenced by strong discriminative power across diverse models and tasks (high AUROC in Table~\ref{tab:public_model_performance}).
However, this emergent capability suffers from significant miscalibration, particularly in RL-trained models.

For practitioners developing LLMs with self-evaluation capabilities, we recommend:
\begin{enumerate}
\item \textbf{Apply post-RL SFT for calibration restoration}: 
Since most modern LLMs use RL as the final training step for performance optimization, a lightweight post-RL SFT phase with self-distillation can restore well-calibrated confidence while preserving task performance gains.
Our experiments demonstrate this approach successfully recovers calibration (ECE: 0.034 in Table~\ref{tab:sft_vs_rl_summary} vs 0.163 baseline in Table~\ref{tab:public_model_performance} for Qwen3) while maintaining accuracy.

\item \textbf{Monitor calibration metrics throughout training}: 
Regularly evaluate both discriminative power (AUROC) and calibration quality (ECE) during training to ensure self-evaluation reliability is not inadvertently degraded in pursuit of task accuracy.
\end{enumerate}

\section{Applications}

Self-evaluation confidence enables effective and efficient detection of potential LLM errors, allowing systems to adaptively invoke correction and risk prevention mechanisms. We demonstrate adaptive RAG as a primary application and discuss other confidence-based applications.

\subsection{Adaptive RAG}
\label{sec:adaptive_rag}

Retrieval-Augmented Generation (RAG) enhances LLM responses by retrieving external context, but retrieval adds latency and cost. We propose adaptive RAG that selectively retrieves only when the model lacks confidence.

We compare two models to evaluate the impact of calibration on adaptive performance: 
\textbf{Instruct} Qwen3-4B-Instruct and Qwen3-4B-SFT from Appendix Table~\ref{tab:qwen3_4b_norm_bins}.

We evaluate on TriviaQA, where questions can be answered either from parametric knowledge (54.21\% accuracy for Qwen3-4B-Instruct) or with retrieved context (72.71\% accuracy). The adaptive system must decide when retrieval is beneficial.

\begin{algorithm}[h]
\caption{Adaptive RAG}
\begin{algorithmic}[1]
\STATE \textbf{Input:} Query $x$, confidence threshold $\tau$
\STATE $y_1, \hat{c}_1 \gets \text{LLM}(x)$\quad // Answer w.o. context
\IF{$\hat{c}_1 < \tau$}
    \STATE $p \gets \text{retrieve}(x)$\quad // Retrieve context
    \STATE $y_2, \hat{c}_2 \gets \text{LLM}(p, x)$\quad // Answer w. ctx
    \RETURN $(y_2, \hat{c}_2)$ if $\hat{c}_2 > \hat{c}_1$ else $(y_1, \hat{c}_1)$
\ELSE
    \RETURN $(y_1, \hat{c}_1)$
\ENDIF
\end{algorithmic}
\end{algorithm}

\textbf{Baselines:} We compare against (1) \textit{No Retrieval}: always answer without context, (2) \textit{Always Retrieve}: always use context.
Table~\ref{tab:adaptive_rag} shows adaptive RAG results on TriviaQA. We make two key observations:

\begin{table}[h]
\centering
\small
\begin{tabular}{lcccc}
\toprule
\textbf{Model} & \textbf{$\tau$} & \textbf{Ret.\%} & \textbf{Gain} & \textbf{Gain/Ret} \\
\midrule
\multirow{5}{*}{\shortstack{Qwen3\\Instruct\\4B}}
 & 0.01 & 21.5\% & +7.86\% & 0.37 \\
 & 0.10 & 25.0\% & +9.09\% & 0.36 \\
 & 0.70 & 29.0\% & +10.24\% & 0.35 \\
 & 0.90 & 30.6\% & +10.62\% & 0.35 \\
 & Always & 100\% & +17.77\% & 0.18 \\
\midrule
\multirow{4}{*}{\shortstack{+SFT}} & 0.10 & 1.6\% & +0.76\% & 0.48 \\
 & 0.20 & 19.4\% & +8.31\% & 0.43 \\
 & 0.70 & 57.6\% & +18.75\% & 0.33 \\
 & 0.90 & 83.5\% & +19.66\% & 0.24 \\
 & Always & 100\% & +19.75\% & 0.20 \\
\bottomrule
\end{tabular}
\caption{Adaptive RAG on TriviaQA. Gain/Ret = retrieval efficiency (accuracy gain per retrieval \%). 
SFT achieves higher efficiency (0.43 vs 0.36) at ~20\% retrieval, showing that better calibration identifies beneficial queries more effectively.}
\label{tab:adaptive_rag}
\end{table}

\textbf{Confidence works out-of-the-box:} The Qwen3-4B-Instruct model achieved +9.09\% accuracy gain by enabling retrieval for only 25\% of queries (at threshold $\tau=0.1$), with retrieval efficiency of 0.36 (gain per retrieval percentage), compared to +17.77\% with always-retrieval (efficiency 0.18). This demonstrated that probability-based confidence provides useful signals for adaptive decision-making even without calibration fine-tuning.

\textbf{SFT enables better control and efficiency:} The SFT model offered two key advantages. 
First, it responded appropriately to threshold changes: retrieval rate ranged from 1.6\% ($\tau=0.1$) to 83.8\% ($\tau=0.8$), 
while the Instruct model remained stuck at ~25-30\% retrieval across $\tau \in [0.1, 0.9]$. 
The sharpened confidence distribution failed to capture graded uncertainty and provided little nuance in confidence estimation (Figure~\ref{fig:calibration_curves}).
Second, at comparable retrieval rates (~20\%), the SFT model achieved higher efficiency (0.43 vs 0.36), demonstrating that better calibration enables more effective identification of which queries truly benefit from retrieval. 
At $\tau=0.7$, the SFT model captured 95\% of maximum gain (+18.75\% out of +19.75\%) with 58\% retrieval.

These results validated our theoretical analysis (Section~\ref{sec:calibration_analysis}): well-calibrated confidence from SFT training enables more effective adaptive decision-making compared to the miscalibrated confidence from RL training.

\subsection{Other Confidence-Based Applications}
\label{sec:other_applications}

Beyond adaptive RAG, confidence-based self-evaluation enables several other efficiency-enhancing applications:

\textbf{Adaptive Reasoning:} For complex reasoning tasks, models can selectively engage expensive Chain-of-Thought reasoning or generate multiple solution attempts only when initial confidence is low while avoiding unnecessary computation for high-confidence queries.

\textbf{Adaptive Escalation:} When confidence falls below a threshold, systems can escalate queries to larger, more capable models. This enables cost-efficient tiered architectures where a small model handles most queries, escalating only uncertain cases to expensive models.

\textbf{Selective Validation:} Confidence thresholds can trigger targeted validation mechanisms—engaging external fact-checking for low-confidence factual claims, or invoking code execution to verify low-confidence program outputs—applying expensive validation only where needed.

\textbf{Human-in-the-Loop:} The AI can selectively seek the user's feedback when confidence is low, enabling efficient human-AI collaboration where human experts focus attention on uncertain cases rather than reviewing all outputs.

\textbf{Iterative Research Termination:} In deep research applications where models need to gather and synthesize information across multiple iterations, confidence scores can determine optimal stopping points to 
prevent both premature termination (missing critical information) and excessive iteration (wasting computational resources).

\section{Future Work}

While RL training optimizes for reward exploitation, this comes at the cost of calibration: responses with even slight advantages receive exponentially more probability mass through iterative updates, eventually approaching near-certainty.
This tension between reward exploitation and calibration suggests two complementary research directions.

\textbf{Calibration-preserving training.}
Future work should explore RL algorithms that preserve calibration—where probabilities reflect the empirical distribution of advantages—while achieving competitive task performance.
This could be achieved through explicit calibration regularization terms or modified advantage weighting schemes.

\textbf{Inference-time exploitation.}
Complementary to calibration-preserving training, inference-time mechanisms can enable reward exploitation while the underlying model maintains calibrated probabilities.
This could be achieved by adjusting the decoding temperature or exploring more advanced beam search algorithms.

Such decoupling of model training and reward exploitation offers two key advantages: (1) the model retains reliable confidence estimates for error detection, and (2) the degree of exploitation can be adjusted at runtime based on task requirements.

\section{Conclusion}
\label{sec:conclusion}

This work establishes that training objectives fundamentally determine confidence reliability: maximum-likelihood estimation naturally produces calibration, while reward optimization induces overconfidence. This insight explains why self-evaluation emerges in modern LLMs yet remains poorly calibrated, and provides a principled solution through post-RL SFT. Our approach enables practitioners to restore reliable uncertainty quantification without sacrificing performance gains from RL training.

Beyond improved metrics, well-calibrated confidence unlocks qualitatively new system behaviors—selective intervention, graceful degradation, and resource-efficient adaptation—that become increasingly critical as LLMs deploy in consequential domains. Future work should develop training algorithms that simultaneously optimize performance and calibration, extend confidence-based adaptation to diverse intervention strategies, and investigate multi-agent collaboration through explicit uncertainty communication.

The path to trustworthy AI demands systems that know what they don't know. By providing theoretical understanding and practical methods for calibrated confidence, this work contributes to responsible AI deployment in high-stakes environments.

\section*{Impact Statement}

This work advances machine learning by enabling LLMs to reliably express uncertainty, with several societal impacts:

\textbf{Positive impacts}: Well-calibrated confidence estimates (1) improve safety by preventing deployment of incorrect outputs in high-stakes domains like healthcare and finance; (2) reduce hallucinations through selective verification mechanisms; (3) enable resource-efficient adaptive systems that apply expensive operations only when needed; and (4) enhance trust by helping users make informed decisions about when to trust, verify, or override model outputs.

\textbf{Limitations and risks}: Our approach requires access to output probabilities, which may not be available in all API deployments. Confidence estimates reflect training data distributions and may be unreliable for out-of-distribution queries or domains with systematic model biases. Users should not rely solely on confidence scores without domain expertise, as miscalibrated confidence—though improved by our methods—can still lead to overconfidence or excessive caution. The approach is most effective when combined with human oversight and other validation mechanisms.

\bibliography{references}
\bibliographystyle{icml2026/icml2026}

\newpage
\appendix
\onecolumn

\section{Self-Evaluation Example}
\label{sec:selfeval_example}

The following example illustrates the self-evaluation process described in Section~\ref{sec:selfeval}.

\noindent\rule{\linewidth}{0.4pt}

\noindent\textbf{User:} Which American-born Sinclair won the Nobel Prize for Literature in 1930?

\noindent\textbf{Assistant:} Sinclair Lewis won the Nobel Prize for Literature in 1930

\noindent\textbf{Self-Evaluation:} Is this answer correct? Answer only Yes/No

\noindent\textbf{Confidence:} Extract $c_s(\text{Yes})$ and $c_s(\text{No})$ from the first token and apply Equation~\ref{eq:selfeval_confidence}

\noindent\rule{\linewidth}{0.4pt}

\section{Normalized Confidence vs. Unnormalized Confidence}

Table~\ref{tab:normalized_vs_unnormalized} compares the discriminative power (AUROC) of normalized confidence (Equation~\ref{eq:confidence}) versus raw unnormalized confidence across five open-source models. Normalized confidence consistently outperforms raw confidence across all models and tasks, demonstrating the importance of accounting for the constrained output space.

For classification tasks, normalization provides substantial improvements, with gains ranging from +2.5\% to +11.0\% AUROC depending on the model and task. The improvements are particularly notable when the model's probability distribution over answer options varies significantly, as seen in AG News and BoolQ for smaller models.

For self-evaluation tasks, normalization provides consistent improvements across all models. GSM8K shows the largest gains (up to +11.6\% for Qwen3-4B), while TriviaQA and TriviaQA-RC show more modest but consistent improvements. The benefit of normalization is evident across diverse model architectures (dense and MoE) and sizes (4B to 30B parameters).

These results validate our use of normalized confidence throughout the paper and support the technical contribution claimed in our comparison with prior work (Section~\ref{sec:related}). Normalization improves discriminative power without affecting calibration quality (ECE remains similar), making it a straightforward enhancement over raw probability approaches.

\begin{table}[h]
\centering
\small
\begin{tabular}{l|ccc|ccc|ccc|ccc|ccc}
\toprule
& \multicolumn{3}{c|}{\textbf{Gemma-3-4B}} & \multicolumn{3}{c|}{\textbf{Gemma-3-12B}} & \multicolumn{3}{c|}{\textbf{GLM-4-9B}} & \multicolumn{3}{c|}{\textbf{Qwen3-4B}} & \multicolumn{3}{c}{\textbf{Qwen3-30B}} \\
\textbf{Task} & Raw & Norm & $\Delta$ & Raw & Norm & $\Delta$ & Raw & Norm & $\Delta$ & Raw & Norm & $\Delta$ & Raw & Norm & $\Delta$ \\
\midrule
BoolQ & .457 & .788 & 331 & .461 & .747 & 286 & .833 & .826 & -7 & .736 & .817 & 81 & .796 & .778 & -18 \\
CQA & .610 & .756 & 146 & .552 & .820 & 268 & .832 & .833 & 1 & .761 & .803 & 42 & .797 & .847 & 50 \\
AG News & .427 & .718 & 291 & .589 & .683 & 94 & .799 & .809 & 10 & .684 & .750 & 66 & .687 & .710 & 23 \\
HellaSwag & .546 & .698 & 152 & .545 & .804 & 259 & .823 & .827 & 4 & .784 & .809 & 25 & .849 & .857 & 8 \\
GSM8K & .687 & .686 & -1 & .800 & .801 & 1 & .733 & .734 & 1 & .657 & .773 & 116 & .783 & .787 & 4 \\
TQA & .520 & .775 & 255 & .806 & .814 & 8 & .749 & .748 & -1 & .810 & .829 & 19 & .846 & .859 & 13 \\
TQA-RC & .516 & .819 & 303 & .827 & .831 & 4 & .771 & .768 & -3 & .856 & .863 & 7 & .835 & .835 & 0 \\
\midrule
\textbf{Avg} & \textbf{.538} & \textbf{.748} & \textbf{210} & \textbf{.654} & \textbf{.771} & \textbf{117} & \textbf{.791} & \textbf{.792} & \textbf{1} & \textbf{.755} & \textbf{.806} & \textbf{51} & \textbf{.799} & \textbf{.810} & \textbf{11} \\
\bottomrule
\end{tabular}
\caption{AUROC comparison of unnormalized (Raw) vs. normalized (Norm) confidence across five open-source models. $\Delta$ shows the gain in per mille (‰). CQA = CommonsenseQA, TQA = TriviaQA. This validates our normalized confidence approach (Equation~\ref{eq:confidence}) as a technical contribution over prior work using raw probabilities.}
\label{tab:normalized_vs_unnormalized}
\end{table}

\section{Benchmark Task Descriptions}
\label{sec:benchmark_details}

\textbf{Classification Tasks:}
\begin{itemize}
\item \textbf{BoolQ}~\cite{clark2019boolq}: Binary classification for Yes/No questions based on passage comprehension.
\item \textbf{AG News}~\cite{zhang2015character}: Article classification into four categories (World, Sports, Business, Science/Technology).
\item \textbf{CommonsenseQA}~\cite{talmor2019commonsenseqa}: Multiple-choice commonsense reasoning with five options (A-E).
\item \textbf{HellaSwag}~\cite{zellers2019hellaswag}: Multiple-choice sentence completion with four options (A-D) requiring commonsense inference.
\end{itemize}

\textbf{Generation Tasks:}
\begin{itemize}
\item \textbf{GSM8K}~\cite{cobbe2021training}: Grade school math word problems requiring multi-step reasoning with numerical answers.
\item \textbf{TriviaQA}~\cite{joshi2017triviaqa}: Question answering with free-form text answers, evaluated in two settings: (1) without context, where models rely solely on parametric knowledge; (2) with context (TriviaQA-RC), where models use up to five retrieved documents (~2k tokens each), which does not guarantee sufficient information, testing both context comprehension and parametric knowledge.
\end{itemize}

\section{Training Data and Hyperparameters}
\label{sec:training_details}

To enable fair comparison across training paradigms, we use identical training data for SFT, RL (GRPO), and DPO, training each for 1,600 steps. This appendix provides complete details for reproducibility.

\subsection{Training Data Construction}

We construct training datasets from 5,000 samples per task (35,000 samples total) drawn from the training splits of seven benchmarks: BoolQ, CommonsenseQA, HellaSwag, AG News, GSM8K, TriviaQA, and their corresponding self-evaluation variants.

\textbf{Classification Tasks.}
For BoolQ, CommonsenseQA, HellaSwag, and AG News, we use the original training labels directly.

\textbf{Generation Tasks with Self-Distillation.}
For GSM8K and TriviaQA, we employ self-distillation to preserve the base model's native reasoning patterns. Direct use of original dataset labels disrupts Qwen3-4B-Instruct's established reasoning style, degrading task performance after fine-tuning.

For GSM8K, we sample multiple reasoning traces from Qwen3-4B-Instruct and select the first trace yielding the correct final answer. 
For TriviaQA, we sample until obtaining an answer that substring-matches one of the provided reference answers. 
This self-distillation approach maintains the model's natural reasoning and answer formatting patterns.

By sampling from the model's own outputs, self-distillation minimizes disruption to the reasoning and response style, effectively preserving the model's behavior and existing capabilities.
When fine-tuning Qwen3-4B using the original GSM8K training dataset, where the reasoning traces differ significantly from Qwen3-4B's intrinsic reasoning patterns, 
we observed a GSM8K accuracy drop from 92.5\% to 83.09\%.
By using self-distilled labels, the GSM8K accuracy increased to 94.62\%, demonstrating the effectiveness of this approach in maintaining model performance while enabling calibrated fine-tuning.

\textbf{Self-Evaluation Data.}
We construct self-evaluation training data by mixing correct and incorrect self-distilled traces for GSM8K and TriviaQA, enabling the model to learn to assess its own answer correctness.

\textbf{Data Formatting.}
All training data are converted to ChatML message format with proper user/assistant conversational turns, appropriate for instruction-tuned models like Qwen3-4B-Instruct.

\subsection{Supervised Fine-Tuning (SFT)}

\textbf{Training objective:} Cross-entropy loss over next-token predictions using ground-truth labels.

\textbf{Hyperparameters:}
\begin{itemize}
\item Optimizer: AdamW with $\beta_1=0.9$, $\beta_2=0.95$, $\epsilon=10^{-8}$
\item Learning rate: $5 \times 10^{-6}$ with linear warmup
\item Batch size: 32 (global batch size across devices)
\item Training steps: 1,600
\item Gradient clipping: Maximum norm of 1.0
\end{itemize}

\subsection{Reinforcement Learning (GRPO)}

\textbf{Training objective:} Group Relative Policy Optimization with advantage-weighted gradients.

\textbf{Reward function:} Binary rewards (1 for correct, 0 for incorrect) based on task-specific verifiable criteria:
\begin{itemize}
\item Classification tasks: Exact match with ground-truth labels
\item GSM8K: Numeric match on final answer
\item TriviaQA: Substring match with reference answers
\item Self-evaluation tasks: Binary correctness of self-assessment
\end{itemize}

\textbf{Hyperparameters:}
\begin{itemize}
\item Base learning rate: $5 \times 10^{-6}$ with linear warmup
\item PPO clip ratio: 0.2
\item KL divergence coefficient: 0.001
\item Training batch size: 32
\item Rollout batch size: 8 samples per prompt
\item PPO mini-batch size: 16
\item Training steps: 1,600
\end{itemize}

\subsection{Direct Preference Optimization (DPO)}

\textbf{Training objective:} Preference-based maximum likelihood optimization over chosen vs. rejected response pairs.

\textbf{Preference pair construction:}
\begin{itemize}
\item \textbf{Chosen responses:} SFT training labels
\item \textbf{Rejected responses:}
  \begin{itemize}
  \item Classification and self-evaluation tasks: Randomly sampled incorrect labels from the available options
  \item GSM8K and TriviaQA: Empty string, representing complete failure to answer
  \end{itemize}
\end{itemize}

\textbf{Hyperparameters:}
\begin{itemize}
\item $\beta$ (temperature parameter): 0.05
\item Learning rate: $2.5 \times 10^{-6}$ with linear warmup
\item Batch size: 32 (16 preference pairs)
\item Training steps: 1,600
\item Gradient clipping: Maximum norm of 1.0
\item Reference model: Frozen copy of Qwen3-4B-Instruct-2507 (pre-trained checkpoint)
\end{itemize}

\section{Detailed Empirical Results For Qwen3-4B}

Table \ref{tab:qwen3_4b_norm_bins} presents the performance of Qwen3-4B-Instruct~\cite{qwen3_2025} along with the model's confidence in its outputs.
We partition samples from each task into 10 equal-mass bins based on the model's output confidence.
For each bin, we calculate the average confidence and average accuracy.
Across all tasks, bins with lower confidence consistently exhibit lower accuracy, 
indicating that the model demonstrates awareness of uncertainty—reflected in its output and self-evaluation logits—when making incorrect predictions.
While model confidence exhibits discriminative power for prediction correctness,

We observe suboptimal calibration between confidence and accuracy across bins for Qwen3-4B-Instruct out of the box.
Confidence values exhibit sharper distributions than accuracy: when accuracy is low, confidence often approaches 0, 
and as accuracy increases, confidence approaches 1 more rapidly than the corresponding accuracy increases.
The same is observed for Qwen3-4B-Instruct model fined-tuned by RL and DPO.
The confidence-accuracy alignment is recovered by SFT with self-distillation.

\begin{table*}[t]
\centering
\scriptsize
\begin{tabular}{c|cc|cc|cc|cc||cc|cc|cc}
\toprule
\textbf{Conf. bin} & \multicolumn{2}{c|}{\textbf{BoolQ}} & \multicolumn{2}{c|}{\textbf{CQA}} & \multicolumn{2}{c|}{\textbf{AG News}} & \multicolumn{2}{c||}{\textbf{HellaSwag}} & \multicolumn{2}{c|}{\textbf{GSM8K}} & \multicolumn{2}{c|}{\textbf{TriviaQA}} & \multicolumn{2}{c}{\textbf{TriviaQA-RC}} \\
Avg. & Acc. & Conf. & Acc. & Conf. & Acc. & Conf. & Acc. & Conf. & Acc. & Conf. & Acc. & Conf. & Acc. & Conf. \\
\midrule
\multicolumn{14}{c}{\textit{Qwen3-4B-Instruct Baseline}} \\
\midrule
1 & 55.05 & 0.939 & 40.98 & 0.792 & 52.76 & 0.884 & 44.72 & 0.793 & 75.57 & 0.934 & 16.50 & 0.000 & 21.57 & 0.000 \\
2 & 64.53 & 1.000 & 51.64 & 0.994 & 71.97 & 1.000 & 54.08 & 0.989 & 83.97 & 1.000 & 20.90 & 0.001 & 31.88 & 0.215 \\
3 & 72.78 & 1.000 & 71.31 & 1.000 & 81.32 & 1.000 & 65.44 & 1.000 & 90.08 & 1.000 & 23.86 & 0.234 & 50.72 & 0.995 \\
4 & 79.51 & 1.000 & 66.39 & 1.000 & 90.92 & 1.000 & 75.90 & 1.000 & 92.37 & 1.000 & 34.34 & 0.977 & 68.62 & 1.000 \\
5 & 88.07 & 1.000 & 81.15 & 1.000 & 91.58 & 1.000 & 84.26 & 1.000 & 93.13 & 1.000 & 45.43 & 1.000 & 85.34 & 1.000 \\
6 & 92.05 & 1.000 & 92.62 & 1.000 & 92.76 & 1.000 & 87.65 & 1.000 & 96.18 & 1.000 & 59.48 & 1.000 & 88.02 & 1.000 \\
7 & 96.33 & 1.000 & 91.80 & 1.000 & 92.76 & 1.000 & 93.23 & 1.000 & 98.47 & 1.000 & 73.80 & 1.000 & 92.98 & 1.000 \\
8 & 97.55 & 1.000 & 90.98 & 1.000 & 92.63 & 1.000 & 95.92 & 1.000 & 97.71 & 1.000 & 82.16 & 1.000 & 94.59 & 1.000 \\
9 & 99.08 & 1.000 & 97.54 & 1.000 & 93.29 & 1.000 & 98.31 & 1.000 & 98.47 & 1.000 & 91.19 & 1.000 & 96.15 & 1.000 \\
10 & 99.69 & 1.000 & 98.37 & 1.000 & 96.05 & 1.000 & 97.02 & 1.000 & 98.57 & 1.000 & 94.33 & 1.000 & 97.16 & 1.000 \\
\midrule
\multicolumn{14}{c}{\textit{Supervised Fine-Tuning}} \\
\midrule
1 & 58.10 & 0.607 & 45.90 & 0.455 & 58.95 & 0.652 & 56.57 & 0.511 & 67.94 & 0.736 & 5.30 & 0.119 & 6.74 & 0.149 \\
2 & 75.23 & 0.755 & 60.66 & 0.614 & 77.63 & 0.844 & 75.60 & 0.687 & 89.31 & 0.893 & 8.14 & 0.169 & 18.62 & 0.273 \\
3 & 84.40 & 0.849 & 68.03 & 0.739 & 86.71 & 0.907 & 90.34 & 0.812 & 95.42 & 0.926 & 15.22 & 0.228 & 51.11 & 0.484 \\
4 & 88.07 & 0.901 & 83.61 & 0.831 & 92.89 & 0.937 & 95.92 & 0.888 & 96.95 & 0.943 & 23.30 & 0.312 & 75.36 & 0.697 \\
5 & 94.80 & 0.933 & 88.52 & 0.898 & 96.58 & 0.955 & 97.51 & 0.932 & 97.71 & 0.951 & 44.04 & 0.457 & 88.02 & 0.804 \\
6 & 97.55 & 0.952 & 96.72 & 0.941 & 97.24 & 0.968 & 98.31 & 0.957 & 98.47 & 0.957 & 63.77 & 0.644 & 92.75 & 0.854 \\
7 & 98.78 & 0.965 & 94.26 & 0.969 & 98.16 & 0.977 & 99.40 & 0.973 & 100.00 & 0.961 & 81.27 & 0.784 & 95.09 & 0.884 \\
8 & 98.17 & 0.974 & 98.36 & 0.984 & 98.82 & 0.984 & 99.80 & 0.983 & 100.00 & 0.965 & 90.80 & 0.860 & 97.71 & 0.903 \\
9 & 98.78 & 0.982 & 99.18 & 0.993 & 97.50 & 0.993 & 99.90 & 0.991 & 100.00 & 0.968 & 95.76 & 0.906 & 98.05 & 0.919 \\
10 & 99.39 & 0.990 & 100.00 & 0.997 & 99.74 & 0.998 & 100.00 & 0.996 & 100.00 & 0.973 & 97.55 & 0.938 & 98.94 & 0.937 \\
\midrule
\multicolumn{14}{c}{\textit{Reinforcement Learning (GRPO)}} \\
\midrule
1 & 59.94 & 0.971 & 48.36 & 0.903 & 61.05 & 0.965 & 50.80 & 0.909 & 73.28 & 0.934 & 14.38 & 0.000 & 17.28 & 0.000 \\
2 & 71.56 & 1.000 & 54.10 & 1.000 & 86.97 & 1.000 & 67.63 & 1.000 & 89.31 & 1.000 & 14.44 & 0.000 & 23.47 & 0.000 \\
3 & 78.29 & 1.000 & 68.03 & 1.000 & 91.58 & 1.000 & 76.69 & 1.000 & 92.37 & 1.000 & 16.72 & 0.000 & 52.90 & 0.015 \\
4 & 85.32 & 1.000 & 81.15 & 1.000 & 93.03 & 1.000 & 87.75 & 1.000 & 97.71 & 1.000 & 30.21 & 0.000 & 76.70 & 0.973 \\
5 & 93.58 & 1.000 & 81.97 & 1.000 & 92.24 & 1.000 & 91.33 & 1.000 & 97.71 & 1.000 & 44.43 & 0.011 & 86.90 & 1.000 \\
6 & 96.33 & 1.000 & 89.34 & 1.000 & 88.82 & 1.000 & 93.92 & 1.000 & 96.18 & 1.000 & 66.56 & 0.959 & 93.14 & 1.000 \\
7 & 95.72 & 1.000 & 94.26 & 1.000 & 86.71 & 1.000 & 94.22 & 1.000 & 99.24 & 1.000 & 79.49 & 1.000 & 94.87 & 1.000 \\
8 & 97.86 & 1.000 & 92.62 & 1.000 & 92.37 & 1.000 & 96.12 & 1.000 & 98.47 & 1.000 & 88.96 & 1.000 & 96.15 & 1.000 \\
9 & 99.39 & 1.000 & 93.44 & 1.000 & 96.45 & 1.000 & 97.61 & 1.000 & 98.47 & 1.000 & 92.70 & 1.000 & 95.71 & 1.000 \\
10 & 99.39 & 1.000 & 98.37 & 1.000 & 98.82 & 1.000 & 98.11 & 1.000 & 98.57 & 1.000 & 96.33 & 1.000 & 96.38 & 1.000 \\
\midrule
\multicolumn{14}{c}{\textit{Direct Preference Optimization (DPO)}} \\
\midrule
1 & 58.10 & 0.969 & 45.08 & 0.900 & 60.13 & 0.966 & 52.29 & 0.935 & 82.44 & 0.845 & 13.94 & 0.000 & 18.00 & 0.000 \\
2 & 75.84 & 1.000 & 62.30 & 1.000 & 78.03 & 1.000 & 72.91 & 1.000 & 94.66 & 1.000 & 16.33 & 0.000 & 21.63 & 0.000 \\
3 & 92.05 & 1.000 & 66.39 & 1.000 & 90.00 & 1.000 & 85.86 & 1.000 & 95.42 & 1.000 & 17.78 & 0.000 & 46.82 & 0.300 \\
4 & 94.80 & 1.000 & 77.05 & 1.000 & 95.53 & 1.000 & 94.42 & 1.000 & 93.13 & 1.000 & 27.09 & 0.000 & 78.54 & 1.000 \\
5 & 93.27 & 1.000 & 90.98 & 1.000 & 95.13 & 1.000 & 98.31 & 1.000 & 95.42 & 1.000 & 41.81 & 0.294 & 94.20 & 1.000 \\
6 & 95.41 & 1.000 & 95.08 & 1.000 & 95.53 & 1.000 & 98.01 & 1.000 & 94.66 & 1.000 & 62.15 & 1.000 & 93.70 & 1.000 \\
7 & 94.80 & 1.000 & 94.26 & 1.000 & 95.79 & 1.000 & 98.01 & 1.000 & 94.66 & 1.000 & 88.46 & 1.000 & 94.70 & 1.000 \\
8 & 96.33 & 1.000 & 94.26 & 1.000 & 94.61 & 1.000 & 98.31 & 1.000 & 93.89 & 1.000 & 86.68 & 1.000 & 93.98 & 1.000 \\
9 & 96.02 & 1.000 & 95.08 & 1.000 & 95.92 & 1.000 & 97.51 & 1.000 & 95.42 & 1.000 & 89.97 & 1.000 & 90.08 & 1.000 \\
10 & 95.11 & 1.000 & 97.56 & 1.000 & 94.61 & 1.000 & 97.91 & 1.000 & 97.86 & 1.000 & 88.49 & 1.000 & 88.49 & 1.000 \\
\bottomrule
\end{tabular}
\caption{Calibration comparison: SFT vs. RL (GRPO) vs. DPO training on Qwen3-4B with identical data. SFT produces well-calibrated confidence through maximum-likelihood estimation; RL and DPO degrade calibration through reward exploitation.}
\label{tab:qwen3_4b_norm_bins}
\end{table*}

\end{document}